\documentclass[10pt,twocolumn,letterpaper]{article}

\usepackage{cvpr}
\usepackage{times}
\usepackage{epsfig}
\usepackage{graphicx}
\usepackage{amsmath}
\usepackage{amssymb}
\usepackage{breqn}
\usepackage{booktabs}
\usepackage{mathtools}
\usepackage{caption}
\usepackage{courier}
\usepackage{xcolor}
\usepackage{microtype}
\usepackage{soul}

\usepackage[pagebackref=true,
  breaklinks=true,
  letterpaper=true,
  colorlinks,
  bookmarks=false,
  linkcolor=red,
  urlcolor=blue,
  citecolor=blue,
  anchorcolor=blue]{hyperref}
  
\DeclareMathOperator*{\argmin}{argmin}
\newcommand{\inv}{\raisebox{.2ex}{$\scriptscriptstyle-1$}}

\cvprfinalcopy %

\def\cvprPaperID{2439} %

\ifcvprfinal\fi
\begin{document}

\title{\papertitle} %

\author{ John Flynn, Michael Broxton, Paul Debevec, Matthew DuVall, Graham Fyffe, \\ 
Ryan Overbeck, Noah Snavely, Richard Tucker\\
{\tt\small \{jflynn,broxton,debevec,matthewduvall,fyffe,rover,snavely,richardt\}@google.com} \\
Google Inc.\\
\vspace*{-1.0cm}
}

\def\thetagt{\theta_{\mathsf{gt}}}
\def\gradcompk[#1]{\mathbf{\nabla}_{n,#1}}
\def\gradcomp{\gradcompk[k]}
\def\ACC{\mathbf{A}}

\def\MPI{\mathbf{M}}
\def\someMPI{M}
\def\MPIH{H}
\def\MPIW{W}
\def\MPID{D}
\def\Image{\mathbf{I}}
\def\CNN{\mathcal{N}_{\omega}}
\def\Dataset{Spaces\xspace}
\def\MPIgt{\MPI_{\mathsf{gt}}}
\def\MPIupk{\MPI^{(k)}}
\def\Imagegt{\Image_{\mathsf{gt}}}
\def\Rgt{\mathcal{R}_{\mathsf{gt}}}
\def\RImagek{\tilde{\mathbf{I}}_k}
\def\papertitle{DeepView: View Synthesis with Learned Gradient Descent}
\newcommand{\citefig}[1]{Fig.~\ref{#1}}
\newcommand{\citeeq}[1]{Eq.~\ref{#1}}
\newcommand{\citetab}[1]{Table~\ref{#1}}
\newcommand{\citesec}[1]{Section~\ref{#1}}

\newcommand{\john}[1]{{\textcolor{red}{[John: #1]}}}
\newcommand{\michael}[1]{{\textcolor{blue}{[Michael: #1]}}}
\newcommand{\noah}[1]{{\textcolor{purple}{[Noah: #1]}}}
\newcommand{\richard}[1]{{\textcolor{orange}{[Richard: #1]}}}
\newcommand{\graham}[1]{{\textcolor{brown}{[Graham: #1]}}}
\newcommand{\ryan}[1]{{\textcolor{green}{[Ryan: #1]}}}
\newcommand{\makevert}[1]{\rotatebox{90}{#1}}

\newcommand{\hlc}[2][pink]{{\sethlcolor{#1}\hl{#2}}}

\twocolumn[{%
\renewcommand\twocolumn[1][]{#1}%
\maketitle
\begin{center}
    \centering
    \includegraphics[width=\textwidth]{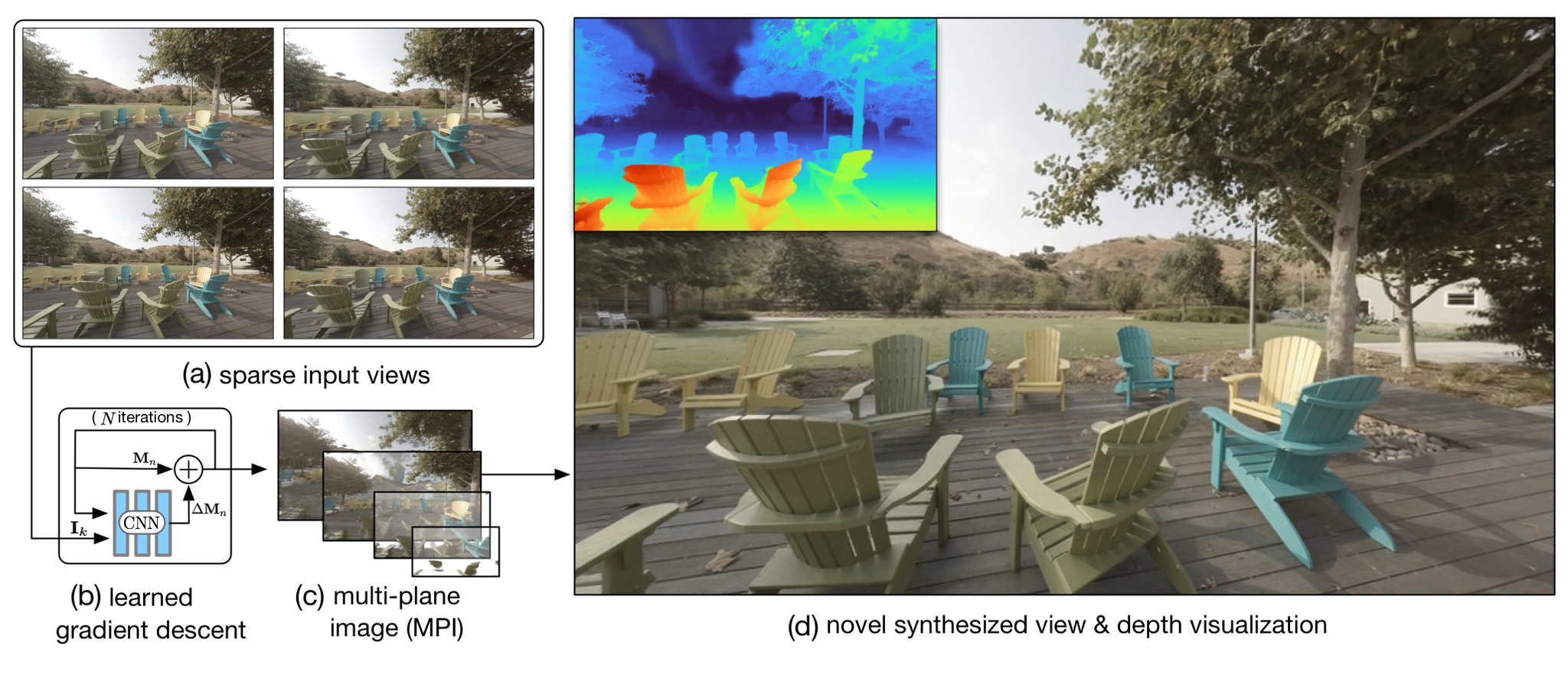}
    \vspace*{-1.0cm}
     \captionof{figure}{{\small The DeepView architecture. {\bf (a)} The network takes a sparse set of input images shot from different viewpoints. {\bf (b, c)} The scene is reconstructed using learned gradient descent, producing a multi-plane image (a series of fronto-parallel, RGBA textured planes). {\bf (d)} The multi-plane image is suitable for real-time, high-quality rendering of novel viewpoints. The result above uses four input views in a $30\text{cm} \times 20\text{cm}$ rectangular layout. The novel view was rendered with a virtual camera positioned at the centroid of the four input views. More results, including video and an interactive viewer, at: \href{https://augmentedperception.github.io/deepview/}{https://augmentedperception.github.io/deepview/}}} 
\label{fig:teaser}
\end{center}
}]
\setlength{\baselineskip}{2.46ex}

\begin{abstract}
We present a novel approach to view synthesis using multiplane images (MPIs). Building on recent advances in learned gradient descent, our algorithm generates an MPI from a set of sparse camera viewpoints. The resulting method incorporates occlusion reasoning, improving performance on challenging scene features such as object boundaries, lighting reflections, thin structures, and scenes with high depth complexity. We show that our method achieves high-quality, state-of-the-art results on two datasets: the Kalantari light field dataset, and a new camera array dataset, \emph{Spaces}, which we make publicly available.
\end{abstract}
\section{Introduction}

Light fields offer a compelling way to view scenes from a continuous, dynamic set of viewpoints. The recently introduced \emph{multiplane image} (MPI) representation approximates a light field as a stack of semi-transparent, colored layers arranged at various depths~\cite{wetzstein:2011:layered,zhou:2018:stereo} allowing real-time synthesis of new views of real scenes. MPIs are a powerful representation that can model complex appearance effects such as transparency and alpha matting. 

Reconstructing an MPI from a sparse set of input views is an ill-posed \textit{inverse problem}, like computed tomography (CT) or image deblurring, where we want to estimate a model whose number of parameters is much larger than the effective number of measurements. Such inverse problems can be solved using gradient descent based optimization methods that solve for the parameters that best predict the measurements via a forward model (such as a renderer in the case of view synthesis). However, for ill-posed problems these methods can overfit the measurements, necessitating the use of priors that are difficult to design and often data dependent. In the case of view synthesis, such overfitting results in synthesized views with noticeable visual artifacts.

Here we present DeepView, a new method for estimating a multiplane image from sparse views that uses \emph{learned gradient descent} (LGD). Our method achieves high quality view synthesis results even on challenging scenes with thin objects or reflections. Building on recent ideas from the optimization community~\cite{adler_primal_dual,adler_inverse_problems,andrychowicz:2016:learning} this method replaces the simple gradient descent update rule with a deep network that generates parameter updates. The use of a learned 
update rule allows for a learned prior on the model parameters through manipulation of the gradients---effectively, the update network learns to generate representations that stay on the manifold of natural scenes. Additionally, the network learns to take larger, parameter-specific steps compared to standard gradient descent, leading to convergence in just a few iterations.

Our method, illustrated in Figure~\ref{fig:teaser}, takes as input a sparse set of views, e.g., from a camera rig (Figure~\ref{fig:teaser}(a)). It then processes the input images using learned gradient descent to produce an MPI (Figure~\ref{fig:teaser}(b,c)). Internally, this module uses a convolution neural network to predict an initial MPI from the input views, then iteratively improves this MPI via learned updates that take into account the current estimate. We show that for this problem the gradients have a particularly intuitive form in that they encode the visibility information between the input views and the MPI layers. By explicitly modeling visibility, our method shows improved performance in traditionally difficult areas such as edges and regions of high depth complexity. The resulting MPI can then be used for real-time view synthesis (Figure~\ref{fig:teaser}(d)).

In addition to the method itself, we also introduce a large, challenging dataset of light field captures called \emph{\Dataset}, suitable for training and testing view synthesis methods. On \emph{\Dataset} as well the standard light field dataset of Kalantari \etal~\cite{kalantari:2016:learning}, we show that our method produces high-quality view synthesis results that outperform recent approaches.

\section{Background and related work}

View synthesis and image-based rendering are classic problems in vision and graphics~\cite{chen:1993:view}. Given a set of input views, one way to pose view synthesis is to reconstruct a light field, a 4D function that directly represents all of the light rays that pass through a volume of space, from which one can generate any view within an appropriate region~\cite{levoy:1996:lightfield,gortler:1996:lumigraph}. However, it is rarely practical to measure a densely sampled light field. Instead, scenes are most often recorded using a limited number of viewpoints that sparsely sample the light field~\cite{rover_lightfields}. View synthesis research has focused on developing priors that aid in recovery of a dense light field from such sparse measurements. 

Some prior view synthesis methods place explicit priors on the light field itself. For example, some degree of smooth view interpolation is made possible by assuming Lambertian reflectance \cite{Levin:2010:linear}, or that the light field is sparse in the Fourier~\cite{shi:2014:lightfield} or Shearlet transform  domain~\cite{vagharshakyan:2018:lightfield}. However, these approaches that do not explicitly model scene geometry can have difficulty correctly distinguishing between a scene's 3D structure and its reflectance properties. Other approaches explicitly reconstruct 3D geometry, such as a global 3D reconstruction~\cite{hedman:2017:casual} or a collection of per-input-view depth maps~\cite{zitnick:2004:interpolation,chaurasia:2013:depth,rover_lightfields} using multi-view stereo algorithms. However, such methods based on explicit 3D reconstruction struggle with complex appearance effects such as specular highlights, transparency, and semi-transparent or thin objects that are difficult to reconstruct or represent.

In contrast, we use a scene representation called the multiplane image (MPI)~\cite{zhou:2018:stereo} that provides a more flexible scene representation than depth maps or triangle meshes. Any method that predicts an MPI from a set of input images needs to consider the visibility between the predicted MPI and the input views so that regions of space that are occluded in some input views are correctly represented.  In this regard, the task is similar to classic work on \emph{voxel coloring}~\cite{seitz:1997:photorealistic}, which computes colored (opaque) voxels with an occlusion-aware algorithm. Similarly, Soft3D~\cite{penner:2017:soft3d} improves an initial set of depth maps by reasoning about their relative occlusion.

One approach to generate an MPI would be to iteratively optimize its parameters with gradient descent, so that when the MPI is rendered it reproduces the input images. By iteratively improving an MPI, such an approach intrinsically models visibility between the MPI and input images. However, given a limited set of input views the MPI parameters are typically underdetermined so simple optimization will lead to overfitting. Hence, special priors or regularization strategies must be used~\cite{wetzstein:2011:layered}, which are difficult to devise.

Alternatively, recent work has demonstrated the effectiveness of deep learning for view synthesis. Earlier methods, such as the so-called DeepStereo method~\cite{flynn:2016:deepstereo} and the work of Kalantari~\etal~\cite{kalantari:2016:learning} require running a deep network for every desired output view, and so are less suitable for real-time view synthesis. In contrast, Zhou~\etal predict an MPI from two input images directly via a learned feed-forward network~\cite{zhou:2018:stereo}. These methods typically pass the input images to the network as a plane sweep volume (PSV) \cite{collins96spacesweep,szeliski99stereomatching} which removes the need to explicitly supply the camera pose, and also allows the network to more efficiently determine correspondences between images. However, such networks have no intrinsic ability to understand visibility between the input views and the predicted MPI, instead relying on the network layers to learn to model such geometric computations. Since distant parts of the scene may occlude each other the number of network connections required to effectively model this visibility can become prohibitively large.

In this work we adopt a hybrid approach combining the two directions---estimation and learning---described above: we model MPI generation as an inverse problem to be solved using a \textit{learned} gradient descent algorithm~\cite{adler_primal_dual,adler_inverse_problems}. At inference time, this algorithm iteratively computes gradients of the current MPI with regard to the input images and processes the gradients with a CNN to generate an updated MPI. This update CNN learns to (1) avoid overfitting, (2) take large steps, thus requiring only a few iterations at inference time, and (3) reason about occlusions without dense connections by leveraging visibility computed in earlier iterations. This optimization can be viewed as a series of gradient descent steps that have been `unrolled' to form a larger network where each step refines the MPI by incorporating the visibility computed in the previous iteration. From another perspective, our network is a recurrent network, that at each step is augmented with useful geometric operations (such as warping and compositing the MPI into different camera views) that help to propagate visibility information.

\section{Method}

We represent a three-dimensional scene using the recently introduced 
multiplane image representation~\cite{zhou:2018:stereo}.  An MPI, $\MPI$, consists of $\MPID$ planes, each with an associated $\MPIH \times \MPIW \times 4$ RGBA image.
The planes are positioned at fixed depths with respect to a virtual reference camera, equally spaced according to inverse depth (disparity), in back-to-front order, $d_1, d_2, ..., d_{\MPID}$, within the view frustum of the reference camera. The reference camera's pose is set to the centroid of the input camera poses, and its field of view is chosen such that it covers the region of space visible to the union of these cameras.
We refer to the RGB color channels of the image for plane $d$ as $c_d$ and the corresponding alpha channel as $\alpha_d$.
To render an MPI to an RGB image $\RImagek$ at a target view $k$ we first 
warp the MPI images and then over-composite \cite{porter:duff:comp} the warped images from back to front:
\begin{equation}
\RImagek = \mathcal{O}(\mathcal{W}_k (\MPI)),
\label{eq:render}
\end{equation}
where the warp operator $\mathcal{W}_k (\MPI)$ warps each of the $\MPID$ MPI images into view $k$'s image space via a homography that is a function of the MPI reference camera, the depth of the plane, and the target view $k$~\cite{zhou:2018:stereo}.
The repeated over operator \cite{porter:duff:comp} $\mathcal{O}$ has a compact form, assuming premultiplied alpha $c_d$\footnote{With
premultiplied color the color channels are assumed to have been multiplied by
$\alpha_{\mathsf{front}}$, so the two-image over operation reduces to $c_{\mathsf{over}} =
c_{\mathsf{front}} + (1-\alpha_{\mathsf{front}})c_{\mathsf{back}}$.}:
\begin{equation}
\mathcal{O}(\someMPI)=\sum_{d=1}^{\MPID} c_d \underbracket[0.5pt]{\prod_{i=d+1}^\MPID(1-\alpha_i)},
\label{eq:over}
\end{equation}
where $\someMPI$ is an MPI warped to a view.
We call the under-bracketed term the \textit{net transmittance} at the depth plane $d$, as it represents the fraction of the color that will remain after attenuation through the planes in front of $d$.

In this paper, we seek to solve the inverse problem associated with Eq.~\ref{eq:render}. That is, we wish to compute an MPI $\MPI$ that not only reproduces the input views, but also generates realistic novel views. 
Since the number of MPI planes is typically larger than the number of input images, the number of variables in an MPI will be much larger than the number of measurements in the input images.

\subsection{Learned gradient descent for view synthesis}

Inverse problems are often solved by minimization, e.g.:
\begin{equation}
\argmin_{\MPI}{\sum_{k=1}^K \mathcal{L}_k(\MPI) + \Phi(\MPI)},
\end{equation}
where
$\mathcal{L}_k(\MPI) = \mathcal{L}(\Image_k, \mathcal{O}(\mathcal{W}_k(\MPI)))$
is a loss function measuring the disagreement between predicted and observed measurements, and $\Phi$ is a prior on $\MPI$.
This non-linear optimization can be solved via iterative methods such as gradient descent, where the update rule (with step size $\lambda$) is given by:
\begin{equation}
\MPI_{n+1} \!= \MPI_{n} - \lambda \left[\sum_{k=1}^K{\frac{\partial\mathcal{L}_k(\MPI_n)}{\partial\MPI_n}} + \frac{\partial\Phi(\MPI_n)}{\partial\MPI_n}\right]\!.
\end{equation}

Recent work on learned gradient descent (LGD)~\cite{adler_primal_dual,adler_inverse_problems} combines classical gradient descent and deep learning by replacing the standard gradient descent rule with a learned component:
\begin{equation}\label{eq:learned-gradient-descent}
\MPI_{n+1} = \MPI_{n} + \CNN \left(
\frac{\partial\mathcal{L}_1(\MPI_n)}{\partial \MPI_n}, 
\dots, 
\frac{\partial\mathcal{L}_K(\MPI_n)}{\partial \MPI_n}, 
\MPI_n \right),
\end{equation}
where $\CNN$ is a deep network parameterized by a set of weights $\omega$. (In practice, as described in Section~\ref{sec:view-synthesis-gradients}, we do not need to explicitly specify $\mathcal{L}$ or compute full gradients.) The network processes the gradients to generate an update to the model's parameters. 
Notice that $\lambda$ and $\Phi$ have been folded into $\CNN$. This enables $\CNN$ to learn a prior on $\MPI$ as well an adaptive, parameter-specific step size.

\begin{figure*}[t]
  \centering \includegraphics[width=\textwidth]{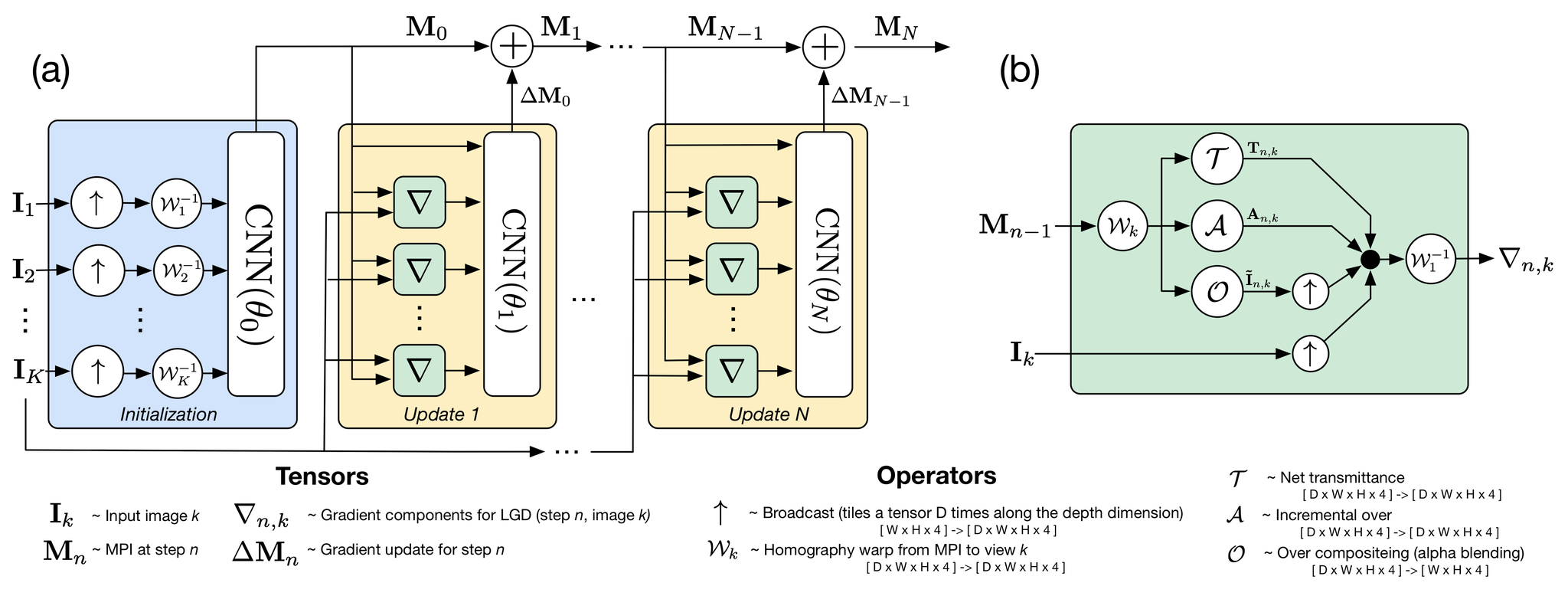}
  \vspace{-0.5cm}
  \caption{\small Our learned gradient descent network. {\bf (a)} An initialization CNN generates an initial MPI based on the plane sweep volumes of the input images. A sequence of update CNNs generate updates to the MPI based on computed gradient components (detailed in {\bf (b)}). All CNNs share the same core architecture (see Fig.~\ref{fig:network}) but are trained with different weights. {\bf (b)} In place of the explicit loss gradient, we compute per-view gradient components, defined in Section~\ref{sec:view-synthesis-gradients}.
  (The black circle represents channel-wise concatenation.)}
  \label{fig:gradient_descent_pt1}
\end{figure*}
To train the network we unroll the $N$-iteration network obtaining the full network, denoted $\mathcal{M}_\omega$. We compute a loss by rendering the final MPI $\MPI_N$ to a held out view and comparing to the corresponding ground truth image $\Imagegt$, under some training loss $\mathcal{L}_f$. The optimization of the network's parameters $\omega$ is performed over many training tuples $(\Image_1,...,\Image_K, \Imagegt)$, across a large variety of scenes, using stochastic gradient descent:
\begin{equation}
\argmin_{\omega} \mathrm{E} \left[ \mathcal{L}_f\left(\Imagegt, \mathcal{O}(\mathcal{W}_{gt}(\mathcal{M}_\omega(\Image_1,...,\Image_K)))\right) \right].
\end{equation}
After training, the resulting network can be applied to new, unseen scenes.

\subsection{View synthesis gradients}\label{sec:view-synthesis-gradients}
LGD requires the gradients of the loss $\mathcal{L}(\RImagek, \Image_k)$ at each iteration. We show that for our problem, for any loss $\mathcal{L}$, the gradients have a simple interpretation as a function of a small set of components that implicitly encode scene visibility. We can pass these gradient components directly into the update networks, avoiding defining the loss explicitly.

The gradient of the $k\textsuperscript{th}$ rendered image $\RImagek=\mathcal{O}(\mathcal{W}_k(\MPI))$ is the combination, through the chain rule, of the gradient of the warping operation $\mathcal{W}_k$ and the gradient of the over operation $\mathcal{O}$. The warp operator's gradient is well approximated\footnote{They are exactly equivalent if the warp is perfectly anti-aliased.} by the inverse warp, $\mathcal{W}^\inv_k$. The gradients of the over operation reduce to a particularly simple form:
\begin{equation}
\frac{\partial \mathcal{O}(\someMPI)}{\partial c_d} = \prod_{i=d+1}^D{(1 - \alpha_i)};
\end{equation}
\begin{equation}
\frac{\partial \mathcal{O}(\someMPI)}{\partial \alpha_d} = \!-\!\Bigg[\sum_{i=1}^{d-1}{c_i}{\prod_{j=i+1}^{d-1}{(1 - \alpha_j)}}\Bigg]\!\!\Bigg[\prod_{i=d+1}^D{(1 - \alpha_i)}\Bigg]\!.
\end{equation}
We refer to the first bracketed term as the \textit{accumulated over} at the depth slice $d$. It represents the repeated over of all depth slices behind the current depth slice. The second bracketed term is again the \textit{net transmittance}. These per slice expressions can be stacked to form a 3D tensor containing the gradient w.r.t.\ all slices. We denote the operator that computes the accumulated over as $\mathcal{A}$ and the corresponding net transmittance operator as $\mathcal{T}$. Define the accumulated over computed in view $k$ as $\mathbf{A}_k=\mathcal{A}(\mathcal{W}_k(\MPI))$ and the corresponding computed net transmittance as $\mathbf{T}_k=\mathcal{T}(\mathcal{W}_k(\MPI))$.

The gradient of \textit{any} loss function w.r.t.\ to $\MPI$ will necessarily be some function of $\RImagek$, $\Image_k$, the accumulated over $\mathbf{A}_k$, and the net transmittance $\mathbf{T}_k$. Thus, without explicitly defining the loss function, we can write its gradient as some function of these inputs:
\begin{equation}
\label{eq:grad_comps}
\frac{\partial \mathcal{L}_k(\MPI)}{\partial \MPI} = \mathcal{F}(
\mathcal{W}^\inv_k(
[\Image_k^\uparrow,
\RImagek^\uparrow,
\mathbf{A}_k,
\mathbf{T}_k]
)),
\end{equation}
where the tensors between the square brackets are channel-wise concatenated and ${}^\uparrow$ represents the broadcast operator that repeats a 2D image to generate a 3D tensor. We define the \textit{gradient components} $\gradcomp = \mathcal{W}^\inv_k([\Image_k^\uparrow, \RImagek^\uparrow, \mathbf{A}_k, \mathbf{T}_k])$ and note that $\mathcal{W}^\inv_k(\Image_k^\uparrow)$ and $\mathcal{W}^\inv_k(\RImagek^\uparrow)$ are the familiar plane sweep volumes of the respective images.

In LGD the computed gradients are passed directly into the network $\mathcal{N}_\omega$; thus $\mathcal{F}$ is redundant---if needed it could be replicated by $\mathcal{N}_\omega$. Instead, we pass the gradient components $\gradcomp$ directly into $\mathcal{N}_\omega$:
\begin{equation}
\MPI_{n+1} = \MPI_{n} + \mathcal{N}_\omega\left(
\gradcompk[1],
\dots,
\gradcompk[K],
\MPI_n\right).
\end{equation}
Explicitly specifying the per-iteration loss is thus unnecessary. (However, we still must define a final training loss $\mathcal{L}_f$, as described in \citesec{subsec:implementation}.) Instead, it is sufficient to provide the network with enough information such that it \textit{could} compute the needed gradients. This flexibility, as shown by our ablation experiments in \citesec{sec:ablation}, allows the network to learn how to best use the rendered and input images during the LGD iterations. We note that this is related to the primal-dual method discussed in~\cite{adler_primal_dual}, where the dual operator is learned in the measurement space.

\begin{figure*}[t]
  \centering \includegraphics[width=\textwidth]{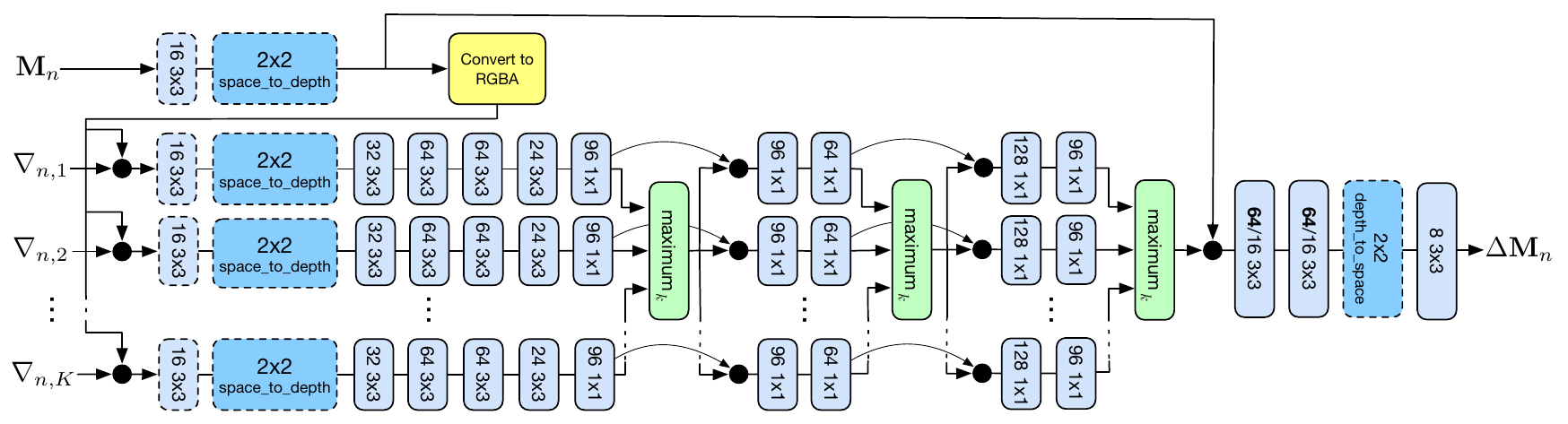}
  \caption{\small The DeepView update CNN. Convolutional layers are labeled with the number of filters followed by the kernel size.
  All convolutions use the Elu \cite{elu} activation, except the last which uses no activation. ``{\bf 64}/16'' indicates the use of 64 features for the initialization step and first iteration and 16 features on all subsequent iterations. Similarly, the downsampling operations with dotted outlines are only used during initialization and the first iteration. The $\textrm{maximum}_k$ operation computes the element-wise maximum across the $k$ input tensors. Black circles represent channel-wise concatenation. The ``Convert to RGBA'' block applies a sigmoid activation to the first four channels of its input to generate valid RGBA values in the range $[0, 1]$, and then premultiplies the RGB channels by alpha. The initialization CNN has the same architecture, however there is no $M_n$ input, and instead of the gradient components we input the PSVs of the input images.}
  \vspace{-0.25cm}
  \label{fig:network}
\end{figure*}

Our learned gradient descent network is visualized in \citefig{fig:gradient_descent_pt1}, and may be intuitively explained as follows.
We initialize the MPI by feeding the plane sweep volumes of the input images through an initialization CNN, similar to the early layers of other view synthesis networks such as Flynn \etal~\cite{flynn:2016:deepstereo}. We then iteratively improve the MPI by computing and feeding the per-view gradient components into an update CNN. Interestingly, initializing using the plane sweep volume of the input images is equivalent to running one update step with the initial MPI slices set to zero color and $\alpha=0$, further motivating the traditional use of plane sweep volumes in stereo and view synthesis.

Intuitively, since the gradient components contain the \textit{net transmittance} and \textit{accumulated over} they are useful inputs to the update CNN as they enable propagation of visibility information between the MPI and the input views. For example if a pixel in an MPI slice is occluded in a given view, as indicated by the value of the net transmittance, then the update network can learn to ignore this input view for that pixel. Similarly, the accumulated over for a view at a particular pixel within an MPI slice indicates how well the view is already explained by what is behind that slice. We note that by iteratively improving the visibility information our method shares similarities with the Soft3D method.

\subsection{Implementation}
\label{subsec:implementation}

We now detail how we implement the described learned gradient descent scheme.

\smallskip
\noindent \textbf{Per-iteration network architecture.}
For both the initialization and update networks we adopt a 2D convolutional neural network (CNN). The input to the CNN is the channel-wise concatenation of the current MPI and the computed gradient components (or, for the first iteration, the PSV of the input images). Within a single iteration the same 2D CNN (with the same parameters) runs on each depth slice, meaning the CNN is fully convolutional in all three of the MPI dimensions and allows changing both the resolution and the number of MPI depth planes after training. This allows performing inference at a high resolution and with a different number of depth planes. 

We adapt recent ideas on aggregating across multiple inputs \cite{pointnet,pointnet_plus,voxelnet} to design our core per-iteration CNN architecture, shown in \citefig{fig:network}. We first concatenate each of the per-view gradient components $\gradcomp$ with the current MPI's RGBA values and transform it through several convolutional layers into a feature space. The per-view features are then processed through several stages that alternate between cross-view max-pooling and further convolutional layers to generate either an initial MPI or an MPI update.
The only interaction between the views is thus through the max pool operation, leading to a network design that is independent of the order of the input views. Further, this network design can be trained with any number of input views in any layout. Intriguingly, this also opens up the possibility of a single network that operates on a variable number of input views with variable layout, although we have not yet explored this.

We use the same core CNN architecture for both the update and initialization CNNs, however, following \cite{adler_primal_dual} we use different parameters for each iteration. As in \cite{adler_primal_dual} we include extra channels ($4$ in our experiments) in addition to the RGBA channels.\footnote{Note that in the discussion below ``MPI'' refers to the network's internal representation of the MPI, which includes these extra channels.} These extra channels are passed from one iteration to the next, potentially allowing the model to emulate higher-order optimization methods such as LBFGS~\cite{lbfgs}. Additionally, we found that using a single level U-Net style network \cite{unet} (i.e.\ down-sampling the CNN input and up-sampling its output) for the first two iterations reduced RAM and execution time with a negligible effect on performance.

\smallskip
A major challenge for the implementation of our network was RAM usage. We reduce the RAM required by discarding activations within each of the per-iteration networks as described in \cite{DBLP:journals/corr/ChenXZG16}, and tiling computation within the MPI volume. Additionally, during training, we only generate enough of the MPI to render a $32 \times 32$ pixel crop in the target image. More details are included in the supplemental material.

\smallskip
\noindent \textbf{Training data.}
Each training tuple consists of a set of input views and a crop within a target view. We provide details of how we generate training tuples from the \emph{\Dataset} dataset in the supplemental material. For the dataset of Kalantari \etal~\cite{kalantari:2016:learning}, we follow the procedure described in their paper.

Our network design allows us to change the number of planes and their depths after training. However, the network may over-fit to the specific inter-plane spacing used during training. We mitigated this over-fitting by applying a random jitter to the depths of the MPI planes during training.

\smallskip
\noindent \textbf{Training loss function.}
We use feature similarity \cite{DBLP:journals/corr/ChenK17aa,unreasonable_feature_loss} as our training loss $\mathcal{L}_f$, specifically the \textit{conv1\_2}, \textit{conv2\_2} and \textit{conv3\_3} layers of a pre-trained VGG-16 network \cite{vgg}, and adopt the per layer scaling method discussed in \cite{DBLP:journals/corr/ChenK17aa}.

\smallskip
\noindent \textbf{Training parameters.}
We implemented our model in TensorFlow \cite{tensorflow2015-whitepaper} and use the ADAM optimizer \cite{ADAM} with learning rate $0.00015$. The supplemental material and \citetab{tab:experiments} further describe the hyperparameters and the training setup used in our experiments.

\section{Evaluation}

We evaluate our method on two datasets: the Lytro dataset of Kalantari~\etal~\cite{kalantari:2016:learning}, and our own \emph{\Dataset} dataset.
The Lytro dataset is commonly used in view synthesis research, but the  baseline of each capture is limited by the small diameter of Lytro Illum's lens aperture. 
In contrast, view synthesis problems that stem from camera array captures are more challenging due to the larger camera separation and sparsely sampled viewpoints. We therefore introduce our new dataset, \emph{\Dataset}, to provide a more challenging shared dataset for future view synthesis research.

\begin{figure}[b]
  \centering \includegraphics[width=\linewidth]{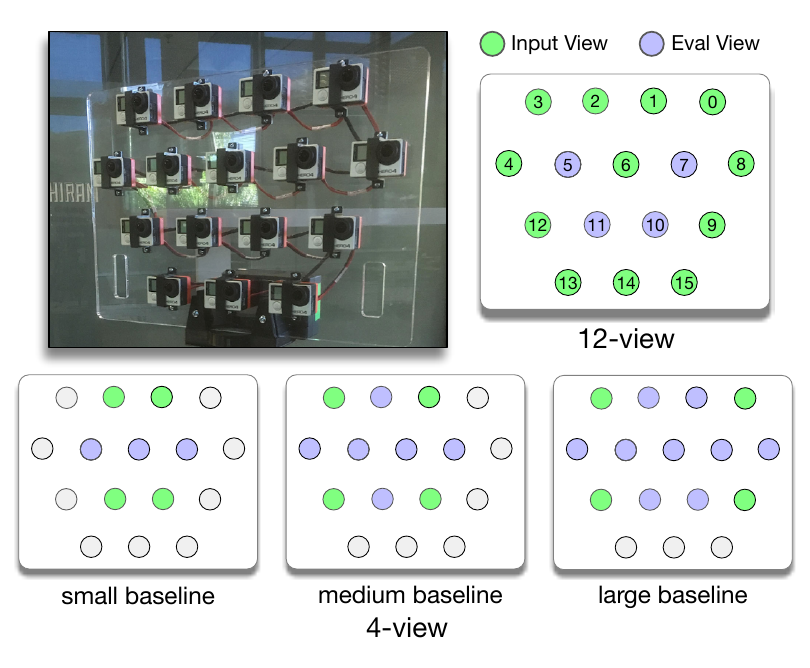}
  \caption{\small The {\em Spaces} dataset was captured using an array of $16$ camera. The horizontal and diagonal distance between cameras is approximately $10$cm. We experimented with different input and evaluation views, as shown. See Sec.~\ref{sec:evaluation} for more details.}
  \label{fig:dataset}
\end{figure}

\emph{\Dataset} consists of 100 indoor and outdoor scenes, captured using a 16-camera rig (see \citefig{fig:dataset}). For each scene, we captured image sets at 5-10 slightly different rig positions (within ${\sim}10$cm of each other). This jittering of the rig position provides a flexible dataset for view synthesis, as we can mix views from different rig positions for the same scene during training.
We calibrated the intrinsics and the relative pose of the rig cameras using a standard structure from motion approach~\cite{hartley_and_zisserman}, using the nominal rig layout as a prior. We corrected exposure differences using a method similar to~\cite{jump}. For our main experiments we undistort the images and downsampl them to a resolution of $800 \times 480$. For our ablation experiments, we used a lower resolution of $512 \times 300$ for expedience. We use 90 scenes from the dataset for training and hold out 10 for evaluation.
\label{sec:evaluation}
\subsection{Quantitative Results}
\newcommand{\cmark}{\ding{51}}%
\newcommand{\xmark}{\ding{55}}%
\begin{table}[tb]
\caption{\small Experimental configurations. A check in the U-Net row indicates that the core CNN ran at a lower resolution for the first two LGD iterations. Other than ``Iterations'', all experiments used 3 LGD iterations. For some experiments we fine-tuned with a larger batch size for the last $20K$ iterations.}
\footnotesize
\centering
\begin{tabular}{@{}l@{\;}c@{\quad}c@{\quad}c@{\;}c@{\;}c@{}}
\toprule
& Kalantari & 4-View & 12-View & Ablation & Iterations \\
\midrule
Input Views & 4 & 4 & 12 & 4 & 4 \\
U-Net & \xmark & \cmark & \cmark & \cmark & \xmark \\
\# Planes training \newline & 28 & 64 & 64 & 20 & 20 \\
\# Planes inference\newline & 28 & 80 & 80 & 20 & 20 \\
Training iterations & 300K & 100K & 100K & 60K & 80K \\
Batch size (fine tune) & 32 (64) & 20 (40) & 20 (40) & 32 & 32 \\
\bottomrule
\end{tabular}
\label{tab:experiments}
\end{table}

We compare our method to Soft3D and to a more direct deep learning approach based on Zhou \etal~\cite{zhou:2018:stereo}. We also ran ablations to study the importance of different components of our model as well as the number of LGD iterations. We provide details on the different experimental setups in \citetab{tab:experiments}. In all experiments we measure image quality by comparing with the ground truth image using SSIM~\cite{ssim}, which ranges between $-1$ and $1$ (higher is better). For the ablation experiments we include the feature loss used during training, which is useful for relative comparisons (lower is better).

\smallskip
\noindent \textbf{Results on the dataset of Kalantari \etal.}
We train a DeepView model on the data from Kalantari \etal~\cite{kalantari:2016:learning} using their described train-test split and evaluation procedure. As shown in \citetab{tab:kalantari_comparison}, our method improves the average SSIM score by 18\% (0.9674 vs.\ 0.9604) over the previous state-of-the-art, Soft3D \cite{penner:2017:soft3d}. 
\begin{table}[tb]
\caption{\small SSIM on the Kalantari Lytro data. (Higher is better.)}
\footnotesize
\centering
\begin{tabular}{@{}l@{\quad}r@{\quad}r@{}c@{\quad\quad}l@{\quad}r@{\quad}r@{}}
\toprule
Scene     & Soft3D & DeepView        & & Scene     & Soft3D & DeepView        \\
\cmidrule{1-3} \cmidrule{5-7}
Flowers1  & 0.9581 & \textbf{0.9668} & & Rock      & 0.9595 & \textbf{0.9669} \\
Flowers2  & 0.9616 & \textbf{0.9700} & & Leaves    & 0.9525 & \textbf{0.9609} \\
Cars      & 0.9705 & \textbf{0.9725} & & (average) & 0.9604 & \textbf{0.9677} \\
\bottomrule
\end{tabular}
\label{tab:kalantari_comparison}
\end{table}

\smallskip
\noindent \textbf{Results on \textit{\Dataset}.}
The \textit{\Dataset} dataset can be evaluated using different input camera configurations. We trained three 4-view networks using input views with varying baselines, and a dense, 12-view network (as shown in \citefig{fig:dataset}). We found that our method's performance improves with the number of input views. When training all networks, target views were selected from all nearby jittered rig positions, although we evaluated only on views from the same rig as the input views, as shown in \citefig{fig:evaluation}. For efficiency in some of our experiments we train on a lower number of planes than we used during inference, as shown in \citefig{fig:evaluation}. The near plane depth for these experiments was set to 0.7m.

\begin{figure*}[t]
  \centering \includegraphics[width=\textwidth]{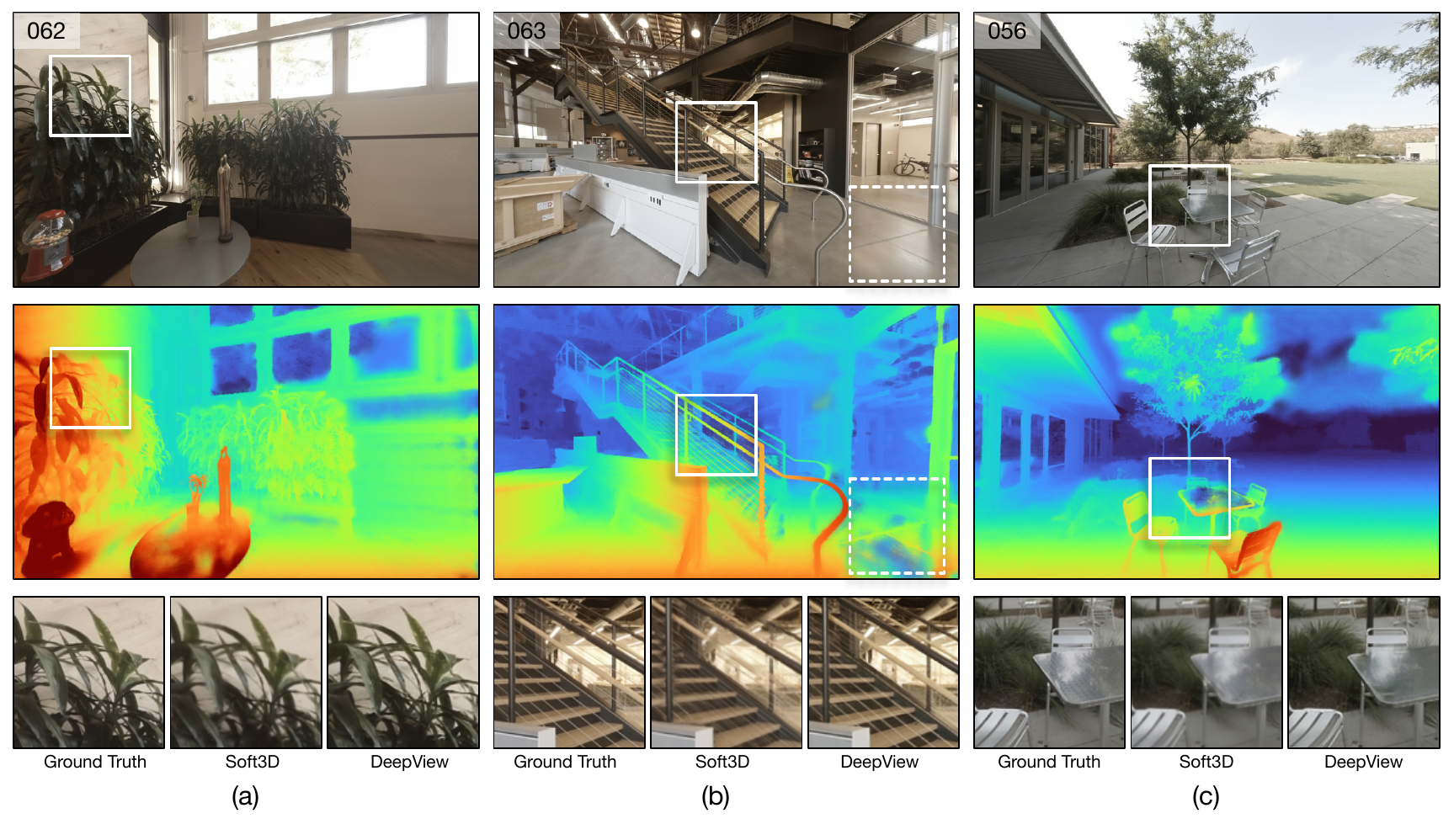}
  \vspace{-0.85cm}
  \caption{\small DeepView results on the Spaces dataset, $12$-view configuration. {\bf (Top row)} Synthesized views using 12 input views shown for evaluation camera viewpoint 7 (cf. Fig.~\ref{fig:dataset}). The top left index denotes the \emph{Spaces} scene number. {\bf (Middle row)} Depth visualizations produced from DeepView MPIs show the recovery of depth information in fine structures such as wires and foliage. {\bf (Bottom row)} Each triplet shows a crop view comparing ground truth (left), Soft3D (center), and DeepView (right). }
  \label{fig:evaluation}
  \vspace{-0.35cm}
\end{figure*}

For comparison, the authors of Soft3D ran their algorithm on {\emph Spaces}. We show these results in \citetab{tab:spaces_comparison}. We also compare DeepView results to those produced with a variant of the network described in \cite{zhou:2018:stereo}. We adapt their network to use 4 wider baseline input views by concatenating the plane sweep volumes of the 3 secondary views onto the primary view (the top left of the 4 views), increasing the number of planes to 40, and increasing the number of internal features by $50\%$. Due to the RAM and speed constraints caused by the dense across-depth-plane connections it was not feasible to increase these parameters further, or to train a 12-view version of this model.

\begin{table}[tb]
\caption{\small SSIM on the \emph{\Dataset} dataset.
\protect{\cite{zhou:2018:stereo}${}^+$} represents the network adapted from \protect{\cite{zhou:2018:stereo}}.}
\footnotesize
\centering
\begin{tabular}{@{}llccr@{}}
\toprule
                         &        &        & \multicolumn{2}{c}{DeepView} \\
Configuration & Soft3D\cite{penner:2017:soft3d} & \cite{zhou:2018:stereo}${}^+$ & 40 Plane & 80 Plane\\
\midrule
4-view (small baseline)  & 0.9260 & 0.8884 &  0.9541  & 0.9561 \\
4-view (medium baseline) & 0.9300 & 0.8874  & 0.9544 & 0.9579 \\
4-view (large baseline)  & 0.9315 & 0.8673  & 0.9485 & 0.9544 \\
12-view                  & 0.9402 & n/a     & 0.9630 & 0.9651 \\
\bottomrule
\end{tabular}
\vspace{-0.5cm}
\label{tab:spaces_comparison}
\end{table}

In all experiments our method yields significantly higher SSIM scores than both Soft3D and the Zhou \etal-based model~\cite{zhou:2018:stereo}. On \emph{\Dataset}, we improve the average SSIM score of Soft3D by 39\% (0.9584 vs.\ 0.9319). Additionally, our method maintains good performance at wider baselines and shows improvement as the number of input images increases from 4 to the 12 cameras used in the 12-view experiments.

The model based on \cite{zhou:2018:stereo} did not perform as well as DeepView, even when both algorithms were configured with the same number of MPI planes (see \citetab{tab:spaces_comparison}). This may because their original model was designed for a narrow-baseline stereo camera pair and relies on network connections to propagate visibility throughout the volume. As the baseline increases, an increasing number of connections are needed to propagate the visibility and performance suffers, as shown by the results for the largest baseline configuration.

\subsection{Ablation and iteration studies}
\label{sec:ablation}
\noindent \textbf{Gradient components.}
In this experiment we set one or more of the gradient components to zero to measure their importance to the algorithm.  We also tested the effect of passing the gradient of an explicit $L_2$ loss during the LGD iterations, instead of the gradient components. Results are shown in \citetab{tab:ablation_comparison}. For experimental details see \citetab{tab:experiments}.

The best performance (in terms of feature loss) is achieved when all gradient components are included. When all components are removed the full network is equivalent to a residual network that operates on each depth slice independently, with no interaction across the depth slices. As expected, in this configuration the model performs poorly.  Between these two extremes the performance decreases, as shown by both the SSIM and feature loss as well as the example images provided in the supplemental material. We note that our training optimizes the feature loss, and the drop in performance as measured by this loss, as opposed to SSIM, is significantly larger.

\smallskip
\noindent \textbf{Number of iterations.}
We also measured the effect of varying the number of LGD iterations from 1 to 4.
A single iteration corresponds to just the initialization network and the performance as expected is poor. The results improve as the number of iterations increases. We note that four iterations does show an improvement over three iterations. 
It would be interesting to test even more iterations. However, due to memory constraints, we were unable to test beyond four iterations, and even four iterations requires too much memory to be practical at higher resolutions. 

\begin{table}[tb]
\caption{\small Ablation study. Each run represents an experiment including the gradient components labeled as
rendered image (\texttt{R}), transmittance (\texttt{T}), and accumulated over (\texttt{A}) with 3 iterations.
The run labeled ``$\nabla$L2'' uses the true gradients of an L2 loss instead of gradient components.
Experiments labeled ``$N$'' use \texttt{RTA} with the indicated number of iterations.
Runs are sorted in order of descending SSIM; feature loss ($\mathcal{L}_f$) (lower is better) is also shown.
}
\footnotesize
\centering
\begin{tabular}{@{}l@{\quad}l@{\quad}l@{}c@{\quad\quad}l@{\quad}l@{\quad}l@{}c@{\quad\quad}l@{\quad}l@{\quad}l@{}}
\toprule
Run & SSIM & $\mathcal{L}_f$ & & Run & SSIM & $\mathcal{L}_f$ & & $N$ & SSIM & $\mathcal{L}_f$ \\
\cmidrule{1-3} \cmidrule{5-7} \cmidrule{9-11}
\texttt{R-A} & 0.9461 & 1.196 & & \texttt{--A} & 0.9397 & 1.271 & & 4 & 0.9461 & 1.146 \\
\texttt{RTA} & 0.9446 & 1.179 & & $\nabla$L2   & 0.9389 & 1.250 & & 3 & 0.9445 & 1.202 \\
\texttt{RT-} & 0.9434 & 1.232 & & \texttt{-T-} & 0.9320 & 1.390 & & 2 & 0.9417 & 1.242 \\
\texttt{-TA} & 0.9435 & 1.238 & & \texttt{---} & 0.9075 & 1.765 & & 1 & 0.8968 & 2.003 \\
\texttt{R--} & 0.9409 & 1.243 & &              &        &       & &   &        &       \\
\bottomrule
\end{tabular}
\vspace{-0.25cm}
\label{tab:ablation_comparison}
\end{table}

\subsection{Qualitative Results}
We visually compare our method to both ground truth and Soft3D in \citefig{fig:evaluation} and in the supplemental material.We notice a general softness in Soft3D results as well as artifacts around edges. In contrast, DeepView produces convincing results in areas that are traditionally difficult for view synthesis including edges, reflections, and areas of high depth complexity. This is best seen in the interactive image comparison tool included in the supplemental material, which allows close examination of the differences between DeepView's and Soft3D's results.

In \citefig{fig:evaluation}a our model produces plausible reflections on the table and convincing leaves on the plant where the depth complexity is high. In \citefig{fig:evaluation}b our method reproduces the fine horizontal railings on the stairs which challenge previous work. \citefig{fig:evaluation}c shows the crisp reconstruction of the complex foliage within the tree. Interestingly, DeepView can even render diffuse reflections, for example in \citefig{fig:evaluation}b (dotted box). The way this is achieved can be seen in the corresponding depth map---transparent alpha values in the MPI permit the viewer to ``see through'' to a reflection that is rendered at a more distant plane.

The crop in \citefig{fig:evaluation}c shows a difficult scene area for both our method and Soft3D. We note that \textit{occluding} specular surfaces are particularly difficult to represent in an MPI. In this example, the MPI places the table surface on the far plane in order to mimic its reflective surface. However, the same surface should also occlude the chair legs which are closer. The end result is that the table surface becomes partially transparent, and the chair legs are visible through it.

Finally, we include a depth visualization in \citefig{fig:evaluation} produced by replacing the MPI's color channels with a false color while retaining the original $\alpha$ values.\footnote{An interactive version of this visualization is included in the supplemental material.} This visualization shows the sharpness of the MPI around edges, even in complex areas such as tree branches.

\section{Discussion}
We have shown that a view synthesis model based on MPIs solved with learned gradient descent produces state-of-the-art results. Our approach can infer an MPI from a set of $4$ input images in approximately 50 seconds on a P100 GPU. The method is flexible and allows for changing the resolution and the number and depth of the depth planes after training. This enables a model trained with medium distance scene objects to perform well even on scenes where more depth planes are needed to capture near objects.

\smallskip
\noindent \textbf{Drawbacks and limitations:}
A drawback of our approach is the complexity of implementation and the RAM requirements and speed of training, which takes several days on multiple GPUs. The MPIs produced by our model share the drawback associated with all plane sweep volume based methods, in that the number of depth planes needs to increase with the maximum disparity. To model larger scenes it may be advantageous to use multiple MPIs and transition between them. Finally, our current implementation can only train models with a fixed number of input views, although our use of max-pooling to aggregate across views suggests the possibility of removing this restriction in the future.

\smallskip
\noindent \textbf{Future work:}
Although our model is not trained to explicitly produce depth we were surprised by the quality of the depth visualizations that our model produces, especially around object edges. However, in smooth scene areas the visualization appears less accurate, as nothing in our training objective guides it towards the correct result in these areas. An interesting direction would be to include a ground truth depth loss during training.

The MPI is very effective at producing realistic synthesized images and has proven amenable to deep learning. However, it is over-parameterized for real scenes that consist of large areas of empty space. Enforcing sparsity on the MPI, or developing a more parsimonious representation with similar qualities, is another interesting area for future work.

\section{Conclusion}

We have presented a new method for inferring a multiplane image scene representation with learned gradient descent. We showed that the resulting algorithm has an intuitive interpretation: the gradient components encode visibility information that enables the network to reason about occlusion. The resulting method exhibits state-of-the-art performance on a difficult, real-world dataset. Our approach shows the promise of learned gradient descent for solving complex, non-linear inverse problems.

\small{\paragraph{Acknowledgments}
We would like to thank Jay Busch and Matt Whalen for designing and building our camera rig, Eric Penner for help with Soft3D comparisons and Oscar Beijbom for some swell ideas.}

{\small
\bibliographystyle{ieee}
\bibliography{learned_gradient_descent}
}

\end{document}


\def\thetagt{\theta_{\mathsf{gt}}}
\def\gradcompk[#1]{\mathbf{\nabla}_{n,#1}}
\def\gradcomp{\gradcompk[k]}
\def\ACC{\mathbf{A}}

\def\MPI{\mathbf{M}}
\def\someMPI{M}
\def\MPIH{H}
\def\MPIW{W}
\def\MPID{D}
\def\Image{\mathbf{I}}
\def\CNN{\mathcal{N}_{\omega}}
\def\Dataset{Spaces\xspace}
\def\MPIgt{\MPI_{\mathsf{gt}}}
\def\MPIupk{\MPI^{(k)}}
\def\Imagegt{\Image_{\mathsf{gt}}}
\def\Rgt{\mathcal{R}_{\mathsf{gt}}}
\def\RImagek{\tilde{\mathbf{I}}_k}
\def\papertitle{DeepView: View Synthesis with Learned Gradient Descent}
\newcommand{\citefig}[1]{Fig.~\ref{#1}}
\newcommand{\citeeq}[1]{Eq.~\ref{#1}}
\newcommand{\citetab}[1]{Table~\ref{#1}}
\newcommand{\citesec}[1]{Section~\ref{#1}}

\newcommand{\john}[1]{{\textcolor{red}{[John: #1]}}}
\newcommand{\michael}[1]{{\textcolor{blue}{[Michael: #1]}}}
\newcommand{\noah}[1]{{\textcolor{purple}{[Noah: #1]}}}
\newcommand{\richard}[1]{{\textcolor{orange}{[Richard: #1]}}}
\newcommand{\graham}[1]{{\textcolor{brown}{[Graham: #1]}}}
\newcommand{\ryan}[1]{{\textcolor{green}{[Ryan: #1]}}}
\newcommand{\makevert}[1]{\rotatebox{90}{#1}}

\newcommand{\hlc}[2][pink]{{\sethlcolor{#1}\hl{#2}}}

\title{DeepView: View synthesis with learned gradient descent \newline Training details} 

\maketitle

\section{Reducing Memory Usage}
A naive implementation of our model requires a prohibitive amount of RAM for both inference and training. Internally, convolutional layers within the CNNs compute activations with as many as 128 channels. The required RAM for the activations for even a single one of these layers would be $K\times D \times N_r \times N_c \times 128$ --- for even moderately sized MPIs this would require hundreds of gigabytes of GPU RAM, and this is for a single network layer within a single iteration of the network. 

Reducing RAM during inference is relatively simple as we can tile the inference of each per-iteration CNN across the  $D\times N_r \times N_c$ MPI volume, recomputing the gradient components for each view as needed.
The tiled inference of the per-iteration CNNs can then be run on GPU. Using this strategy inference takes $50$s for a $12$ input $64 \times 980 \times 580$ MPI on a P100 GPU.

Reducing RAM at training time is more complex. The inference method described above is not directly applicable since we need to back-propagate the gradients. More importantly, a deep network typically requires tens or hundreds of thousands of iterations to converge and a per step time of $50s$ would make the training time impractical. Instead, during training the network is trained to produce a small $32 \times 32$ crop within a target image. By carefully considering both the CNN padding and the MPI and view geometry at each iteration we can compute both the needed crops from each input view and the minimal volume of the MPI that needs to be computed at each iteration in order to produce a given target image patch without border effects. 
Note that this calculation is complex since in order to produce the gradient components for a specific input view crop at iteration $n$ we need to have available the MPI volume visible to that crop at the previous iteration $n-1$, but in order to compute \b\textbf{that} MPI volume in $n-1$ we need to have more of the input view's area etc. The required MPI volume that needs to computed for a given target crop thus increases as the number of LGD iterations increases.

To further reduce RAM we discard activations within each of the per-iteration networks, as described in \cite{DBLP:journals/corr/ChenXZG16} and tile computation along the depth plane dimension.
Even with these memory optimizations, at higher resolutions we are limited to training a single example patch at a time and rely on synchronized replicas to get large effective batch sizes.

\section{Generating training samples from the \Dataset dataset}
To generate a sample we first randomly select a scene and a random rig position. We then randomly select the input views from the set of possible view sets within the rig (see Figure 4). These views form the input to the network. The target view is then randomly selected from the remaining views across all rigs that are within $6$cm of of the convex hull of the input views, and a maximum of $7$cm from the plane of the input views. A $32 \times 32$ crop is then chosen from the target view.

\section{Training hyperparameters}
We used distributed training with synchronized replicas to increase the effective batch size. We typically used $16$ replicas, but for some experiments we used less replicas but computed gradients from two examples in each replica before accumulating the batch.

As in Adler \etal ~\cite{adler_primal_dual} we used global gradient clipping, with a threshold of 8.0.

We use feature similarity \cite{DBLP:journals/corr/ChenK17aa,unreasonable_feature_loss} as our training loss $\mathcal{L}_f$, specifically the \textit{conv1\_2}, \textit{conv2\_2} and \textit{conv3\_3} layers of a pre-trained VGG-16 network \cite{vgg}. The \textit{conv3\_3} layer has a receptive field of $40 \times 40$, so we first reflect padded the $32 \times 32$ target and training crop to be $40 \times 40$. Following Chen \etal \cite{DBLP:journals/corr/ChenK17aa} we then computed the loss by summing the $L_1$ difference of the layers, weighted by the empirically determined weights of $[11.17, 35.04, 29.09]$. Finally, we divided the loss by the area of the crop $32 \times 32$, producing a loss in a more reasonable range which improved the numerical stability of global gradient clipping.

{\small
\bibliographystyle{ieee}
\bibliography{learned_gradient_descent}
}